\documentclass[conference]{IEEEtran}
\IEEEoverridecommandlockouts
\usepackage{cite}
\usepackage{amsmath,amssymb,amsfonts}
\usepackage{algorithmic}
\usepackage{graphicx}
\usepackage{textcomp}
\usepackage{xcolor}
\usepackage{times}
\usepackage{latexsym}
\usepackage{wrapfig}
\usepackage[T1]{fontenc}
\usepackage[utf8]{inputenc}
\usepackage{CJKutf8}
\usepackage{microtype}
\usepackage{booktabs}
\usepackage{hyperref}
\usepackage{multirow}
\def\BibTeX{{\rm B\kern-.05em{\sc i\kern-.025em b}\kern-.08em
    T\kern-.1667em\lower.7ex\hbox{E}\kern-.125emX}}
\begin{document}

\makeatletter
\newcommand{\linkbreakand}{% 
\end{@IEEEauthorhalign}
\hfill\mbox{}\par
\mbox{}\hfill\begin{@IEEEauthorhalign}
}
\makeatother
\title{Prompt-Learning for Cross-Lingual Relation Extraction}

% \author{\IEEEauthorblockN{Anonymous Authors}}

% \author{\IEEEauthorblockN{
%         \thanks{*Work was done when C. Hsu and C. Zan were interning at JD Explore Academy.} 
%         Chiaming Hsu\thanks{C. Hsu, F. Lin, and W. Hu are with Wuhan University (e-mail:2020262110001@whu.edu.cn, linfu@whu.edu.cn, hwb@whu.edu.cn).},
%         Changtong Zan\thanks{C. Zan and W. Liu are with China University of Petroleum (East China) (e-mail: b20050011@s.upc.edu.cn; liuwf@upc.edu.cn).},
%         Liang Ding$^\ddagger$\thanks{L. Ding is with the JD Explore Academy at JD.com (e-mail: dingliang1@jd.com).}, 
%         \\
%         Longyue Wang\thanks{L. Wang is with the Tencent AI Lab (e-mail: vinnylywang@tencent.com)},
%         Xiaoting Wang\thanks{X. Wang is with JD.com (e-mail: wangxiaoting35@jd.com)},
%         Weifeng Liu, 
%         Fu Lin$^\ddagger$, 
%         Wenbin Hu$^\ddagger$,
%         \thanks{$^\ddagger$Corresponding Authors: Wenbin Hu (e-mail: hwb@whu.edu.cn), 
%         Fu Lin (e-mail: linfu@whu.edu.cn), 
%         Liang Ding (e-mail: dingliang1@jd.com).}
%         }}

\author{\IEEEauthorblockN{
        \thanks{*Work was done when C. Hsu and C. Zan were interning at JD Explore Academy.} 
        Chiaming Hsu\thanks{C. Hsu, F. Lin, and W. Hu are with Wuhan University.},
        Changtong Zan\thanks{C. Zan and W. Liu are with China University of Petroleum (East China).},
        Liang Ding$^\ddagger$\thanks{L. Ding is with the JD Explore Academy at JD.com.}, 
        \\
        Longyue Wang\thanks{L. Wang is with the Tencent AI Lab.},
        Xiaoting Wang\thanks{X. Wang is with JD.com.},
        Weifeng Liu, 
        Fu Lin$^\ddagger$, 
        Wenbin Hu$^\ddagger$,
        \thanks{$^\ddagger$Corresponding Authors: Wenbin Hu (e-mail: hwb@whu.edu.cn), 
        Fu Lin (e-mail: linfu@whu.edu.cn), 
        Liang Ding (e-mail: dingliang1@jd.com).}
        }}
\maketitle

\begin{abstract}
Relation Extraction (RE) is a crucial task in Information Extraction, which entails predicting relationships between entities within a given sentence. However, extending pre-trained RE models to other languages is challenging, particularly in real-world scenarios where Cross-Lingual Relation Extraction (XRE) is required. Despite recent advancements in Prompt-Learning, which involves transferring knowledge from Multilingual Pre-trained Language Models (PLMs) to diverse downstream tasks, there is limited research on the effective use of multilingual PLMs with prompts to improve XRE. In this paper, we present a novel XRE algorithm based on Prompt-Tuning, referred to as Prompt-XRE. To evaluate its effectiveness, we design and implement several prompt templates, including hard, soft, and hybrid prompts, and empirically test their performance on competitive multilingual PLMs, specifically mBART. Our extensive experiments, conducted on the low-resource ACE05 benchmark across multiple languages, demonstrate that our Prompt-XRE algorithm significantly outperforms both vanilla multilingual PLMs and other existing models, achieving state-of-the-art performance in XRE. To further show the generalization of our Prompt-XRE on larger data scales, we construct and release a new XRE dataset- WMT17-EnZh XRE, containing 0.9M English-Chinese pairs extracted from WMT 2017 parallel corpus. Experiments on WMT17-EnZh XRE also show the effectiveness of our Prompt-XRE against other competitive baselines. The code and newly constructed dataset are freely available at \url{https://github.com/HSU-CHIA-MING/Prompt-XRE}.
\end{abstract}


\section{Introduction}
\label{sec:introduction}
The task of \textit{Information Extraction (IE)} involves the automatic extraction of structured data from unstructured or semi-structured texts. This process enables the utilization of knowledge contained in text material in the creation of knowledge graphs or relational databases, which have numerous applications in various areas, including recommender systems~\cite{2020Improving}, fine-grained sentiment analysis~\cite{Wang2022ACC}, question answering~\cite{2021QA}, and machine translation~\cite{ding2021progressive}.

\textit{Relation Extraction (RE)} is a fundamental task in the IE field, aimed at categorizing the connections between entities in a text or the semantic connections in an event~\cite{doddington-etal-2004-automatic}. This task is crucial for several NLP applications.
For instance, the relationship between entities such as``\textit{Steve Jobs}'' and ``\textit{Apple Inc.}'' can be extracted as ``\textit{founder}'', which can then be used to answer questions like ``\textit{Who is the founder of Apple Inc.?}''  However, conventional RE methods often require a large amount of supervised training data, which can be costly to obtain, and may also struggle with new relations.
Recently, zero-shot setting RE task has attracted the attention of the community, as it could generalize to unseen relation sets with no requirement of annotated data~\cite{10.1145/3293318}.

RE originated from feature-based statistical models, e.g.~\cite{miwa-bansal-2016-end}, and recent works mainly focus on one single language, e.g. English.
However, there are thousands of languages, and we need a cross-lingual relation extraction (XRE) model to extract relations for different languages. Also, as the annotation of relation with entities is difficult to obtain when extends to more languages, it urges us to focus on the zero-shot XRE setting, which predicts the relation for the target language based on a model trained on another language. For example,~\cite{ahmad2021gate} use universal dependency parse of sentences to add dependency synthetic information to the graph neural network during training, has achieved good performance in XRE in the \textbf{ACE05} dataset which has shown the effectiveness for XRE task, but this can be cumbersome when processing data and faces the problem of high overhead during model training. 

Pre-trained multilingual Language Models (PLM) can be successfully applied to a variety of downstream tasks, for example, translation~\cite{liu2020multilingual}, Named Entity Recognition~\cite{yan2021unified}.
In situations where the labeled data is insufficient, prompt-tuning is useful for incorporating PLM knowledge to enhance the outcomes of downstream operations~\cite{brown2020language}. The key to prompt-tuning is to carefully craft the task input structure such that it closely resembles the pre-trained PLMs procedures and more effectively fosters PLMs expertise.
Previously, most work on prompt-based tuning~\cite{zhang2021differentiable} mainly considers monolingual prompts. Multilingual activities often face difficulties due to a shortage of multilingual prompts and the dependence on native language specialists for the creation of templates and tagging terms. One potential solution to this issue is translating the source prompts into the target language. However, the accuracy of these translations is uncertain, particularly for languages with limited resources.
Alternative approaches have explored the use of \textbf{soft prompts} comprised of continuous vectors. While this method avoids the challenge of generating prompts for multiple languages, it presents its own difficulties as there is a mismatch between pre-training and prompt-tuning methods, as the \textbf{soft prompt} is not utilized in the pre-training phase of the model.
Therefore, there remains a significant challenge in effectively applying multilingual prompts to multilingual activities, and further research is needed to address the limitations of current methods.

In this work, we present a zero-shot transfer approach for cross-lingual relation extraction (XRE) based on the prompt-tuning of a pre-trained multilingual language model. Our approach effectively leverages datasets from different languages for XRE. Specifically, the model is first trained on one language through the zero-shot transfer method, and then further improved through prompt-tuning for the relation extraction task. Furthermore, we propose a flexible prompt template design that is applicable to a seq2seq model and not limited to a specific language or dataset. we evaluated the effectiveness of our proposed zero-shot transfer XRE approach through experiments on the large-scale \textbf{ACE05} dataset~\cite{2006ACE}. Due to the scarcity of parallel corpus-based XRE datasets, we have created a new dataset for XRE tasks based on the \textbf{WMT17-EnZh XRE} dataset. The construction of this new dataset is explained in detail in Section \ref{sec:wmtenzh}. The results demonstrate that our approach delivers promising outcomes.
% Additionally, we recognize the importance of a parallel corpus-based XRE dataset and therefore extracted data from the \textbf{WMT17-EnZh XRE} dataset to annotate relations for XRE. 
Also, the evaluation of our model in the non-zero-shot setting confirms its robustness and efficacy.
The main contributions of the paper include:
\begin{itemize}
\item We design a prompt-learning strategy for cross-lingual relation extraction (XRE) task, namely \textsc{\bf Prompt-XRE}, upon multilingual pre-trained Multilingual language models, e.g. mBART, allowing the model to achieve better zero-shot learning ability. To the best of our knowledge, this is the first work to borrow the success of prompt learning into the XRE task. 
\item 
We conduct extensive experiments upon widely used benchmarks spanning different languages to validate the effectiveness of our deliberately designed prompt templates, achieving the state-of-the-art XRE performance on the \textit{low-resource} {ACE05} dataset.
\item 
To address the data scarcity problem of the XRE task, we construct and release a 0.9M English$\rightarrow$Chinese dataset, namely WMT17-EnZh XRE, with WMT parallel corpus. A \textit{large dataset} of experiments also demonstrates the efficacy and applicability of our prompt-learning strategy.  

\end{itemize}

\section{Task Description}
\label{sec:task}
\subsection{Zero-Shot Cross-Lingual Relation Extraction}
% \paragraph{\bf Relation Extraction (RE)}: 
Relation Extraction (RE) is the task of identifying the types of relationships of ordered entity mention pairs ~\cite{doddington-etal-2004-automatic}. For example, the relationship between ``Terrorists'' (entity1) and ``hotel'' (entity2) in the sentence ``Terrorists started firing at the hotel.'' is \textbf{PHYS: Located}.
\paragraph{\bf Zero-shot Cross-Lingual Relation Extraction}
Zero-shot XRE describes a situation in which the target language lacks any labeled examples. Before deploying the models in target languages, we train XRE models on a single (single-source) language.

\section{Background}
\label{sec:back}
\subsection{Generative Relation Extraction}
Traditional restrict domain RE is usually based on a small vertical corpus and pre-defined target relations, and when migrating to new domain data, the relationship classes and rules for relationship extraction need to be redefined. So this approach is domain dependent. Some of the work now focuses on a simpler syntactic task: open-ended RE, which does not require pre-defined relationship types but extracts entities and their relationships directly from the unstructured text. However, the semantics of the relational information extracted by this approach is not clear, so it is also difficult to be applied to downstream tasks without aligning it with the relevant knowledge base~\cite{broscheit2017openie}.

In this work, we propose the use of a Seq2Seq network structure for Cross-Lingual Relation Extraction (XRE). Seq2Seq, a deep learning-based architecture, operates by encoding the original information to obtain the semantic representation of the full text. This encoded representation is then used to decode the information step by step, leading to the generation of the final output. The use of Seq2Seq in the XRE task allows us to leverage the capabilities of deep learning to extract relationships between entities in a given sentence, even in cross-lingual scenarios where the language may be different from the training data. This approach is expected to significantly improve the performance of XRE, as compared to traditional methods.

\subsection{Prompt Learning}
The prompt learning approach has achieved significant advancements in various NLP tasks due to the creation of the GPT-3 language model. Several studies have demonstrated the effectiveness of manual prompts in optimizing the prompt learning process~\cite{zhong2023chat,Lu2023EAPrompt,Peng2023ChatGPT4MT}. The verbalizer has been calibrated to incorporate external knowledge, as proposed in~\cite{hu2021knowledgeable}. Automated searches for discrete prompts have gained popularity as they eliminate the need for manual prompt manufacturing~\cite{gao2020making}. The study in~\cite{shin2020autoprompt} suggested gradient-guided searches to create templates and automatically tag words in the vocabulary. Recently, continuous embeddings that can be learned, as opposed to labeled words, have been proposed as templates for continuous prompts~\cite{li2021prefix}.

In this paper, we have designed several prompt templates for XRE tasks, including the following three types:
\begin{itemize}
\item \textbf{Hard Prompt}: Manual design of prompts using natural language.
\item \textbf{Soft Prompt}: Use artificial tokens as templates rather than discrete tokens.
\item \textbf{Hard-Soft Prompt}: Prompts designed by combining natural language with manually designed non-discrete tokens.

\end{itemize}
\section{Methodology}
\label{sec:meth}
In this study, we aim to tackle the challenge of zero-shot cross-lingual relationship extraction (XRE) by proposing a novel approach based on prompt-tuning of a pre-trained multilingual language model. The zero-shot XRE task requires training the model on source language A and directly evaluating its performance on target language B. To achieve this, we first introduce the task formulation and construct a prompt template that can be applied to the seq2seq model. Subsequently, we describe the utilization of the mBART model and present its formal formulation. In the following sections, we will elaborate on the implementation details and experimental results of our approach, demonstrating its effectiveness in exploiting the cross-lingual information for the zero-shot XRE task.

\subsection{Preliminary}
\paragraph{\bf Denoising Task for Multi-Lingual Sequence-to-Sequence Pretraining}
The denoising pretrain task learns a function that directly maps the sentence with noises to the original sentence. We adopt mBART~\cite{tacl_mBART} as the backbone model, which uses a standard sequence-to-sequence Transformer~\cite{liu2021understanding}, pre-trained on the CC-25\footnote{\url{https://github.com/pytorch/fairseq/tree/main/examples/mbart}} dataset. For a sample $x^l$ in language $l$, mBART is trained to capture the monolingual knowledge during the pretraining stage by optimizing follow objective: 
\begin{eqnarray}
\label{eq:denoise}
\mathrm{F}_\mathrm{PT}=\sum_{l\in L}-\log P(x^l|N(x^l)),
\end{eqnarray}
where $L$ represents the set of languages contained in CC-25, $\mathrm{N}(*)$ is the noise function containing span masking and sentence permutation. The former is more important to help mBART improve the finetuning performance on sentence-level generation tasks. As set in the article~\cite{lewis2019bart}, $35\%$ of all words are chosen for span masking, and each span is changed to a mask token. A Poisson distribution with an expected length of 3.5 determines the span length.

\paragraph{\bf Standard Fine-tuning for Cross-Lingual Tasks}
Given a sentence pair $ \textit{X} $, the standard fine-tuning directly feed the source sentence into the encoder and decodes the target sentence. 
Following mBART, we inject the language identification signal. 
During decoding, the language id ``[ZH]'' is generally attached to the end of the input sequence and ``[EN]'' is prepended to the beginning of the output sequence.
Specifically, we append the source language id to the end of the input sequence, and target language id is prepended to the beginning of the output sequence during decoding. 
We also add a symbol ``</s>'' to mark the end of each sentence. However, since our method is zero-shot cross-lingual transfer, the language id add to our input are all the same. For example, the language id added when training the model in English are all ``[EN]'', and the language id added when the model is trained and tested in Chinese are all ``[ZH]''. 

\subsection{Prompt-Based Fine-Tuning for Zero-Shot Transfer XRE}
\paragraph{\bf Relation Extraction Task} An RE task can be denoted as d = [\textit{X}, \textit{Y}], where \textit{Y} is the set of relation labels in the target language and \textit{X} is the set of instances in the source language. For each instance \textit{x} = \{${e_1 , e_2, e_3, e_s,...., e_o, ....., e_n}$\}. Since one entity may have numerous tokens, we just use $e_s$ and $e_o$ to represent all entities since the purpose of RE is to predict the relation $y \in Y$ between subject entity $e_s$ and object entity $e_o$.
%%%%%%%%%%%%%%%%%%%%%%%%%%%%%%%%%%%%%%%%%%%%%%
\begin{figure}[]  
\centering  
\includegraphics[height=3.0cm, width=9.0cm]{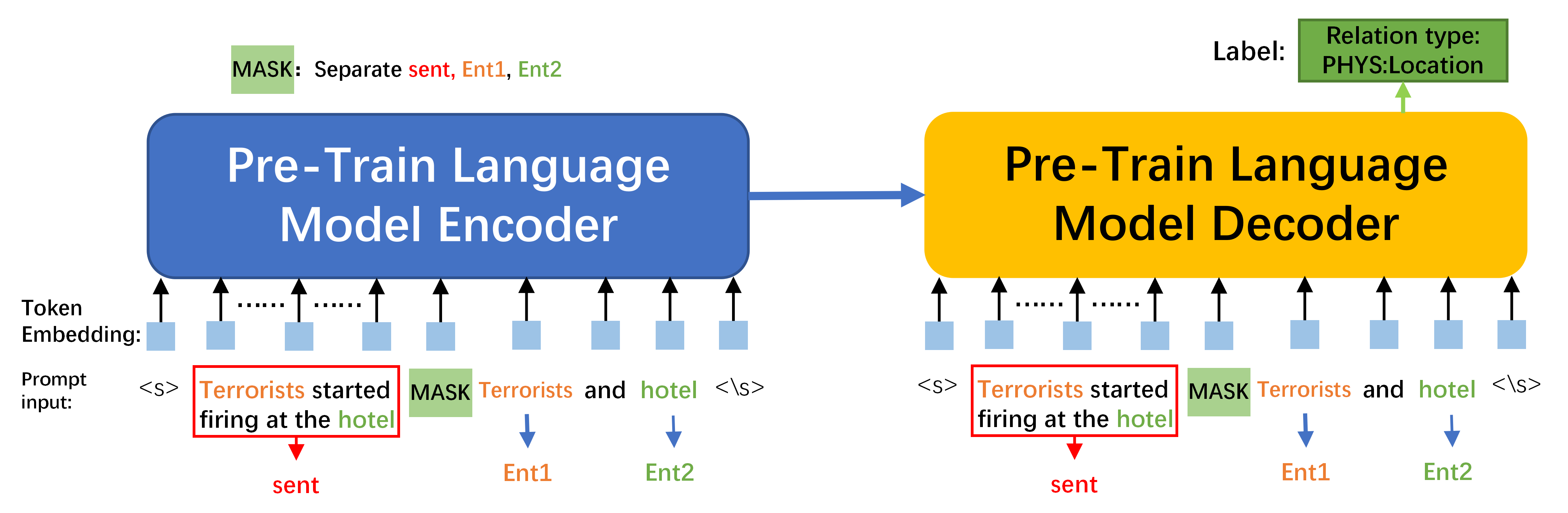}
\caption{\textbf{Prompt-tuning pre-trained language Model.} The above encoder input and decoder input are both in the same language.}
\label{figure1}
\end{figure}
%%%%%%%%%%%%%%%%%%%%%%%%%%%%%%%%%%%%%%%%%%%%%%

\paragraph{\bf Fine-Tuning for Relation Extraction}
Given a PLM \textit{M} for RE. Earlier fine-tuning approach first transform the instance $\textit{x}$ = $\{{e_1 , e_2, e_3, e_s,...., e_o, ....., e_n}\}$, into a PLM input sequence, such as ``<s> \textit{x} </s>''. The output hidden vectors of the \textit{M} encode the input sequence as $\textbf{v}_{enc}$ = $\{{\bf{v_{<s>} , v_1 , v_2 , v_s ,...., v_o , ....., v_{</s>}}}\}$. Then input $\textbf{v}_{enc}$ and ``<s>  $\textit{x}$ </s>'' into the decoder of model \textit{M} to get $\textbf{v}_{dec}$. The following formula denotes the conditional probability of the relation type $y$: $\textit{p}(y|x) = Softmax(\textbf{Wv}_{dec})$.

\paragraph{\bf Prompt-Tuning of Pre-Trained Multilingual Language Model}
Prompt tuning is suggested as a way to close the gap between upstream activities and pre-training. The most difficult task was creating a proper label V and template $t(\cdot)$, which are referred to jointly as a prompt P.
The template maps each occurrence of x to an input prompt, $x_{prompt}=T(x)$. The template $t(\cdot)$ specifically refers to where and how many extra words are added. V denotes a collection of tag words in the vocabulary of the model $\textit{M}$, where $MP: \Upsilon \rightarrow V$ is an injective mapping that links label words V to task labels.
In the traditional method, in addition to preserving the original tags of $x_{prompt}$, one or more [MASK] are placed to the boundary of $\textit{M}$ to fill in the tag words. However, our approach uses [MASK] as a token to hint at the model $\textit{M}$ and is used to segment the model input in entities or sentences. Since $\textit{M}$ can predict the correct label, we are able to formalize $p(y|x)$ terms of the probability distribution of V at the masked position, that is, $p(y|x)$=$p(MP(y)|x_{prompt})$. Figure \ref{figure1} depicts our prompt-tuning PLMs for categorizing based on the mBART model.

\paragraph{\bf Zero-Shot Cross-Lingual Transfer Relation Extraction}
We train XRE models on a single language (single-source) before deploying them in target languages. As seen in Figure \ref{figure2}, we train a XRE model using English as the source language, test the XRE model using Chinese as the target language. 

%%%%%%%%%%%%%%%%%%%%%%%%%%%%%%%%%%%%%%%%%%
\begin{figure}[htp]  
\centering  
\includegraphics[height=4.5cm, width=9cm]{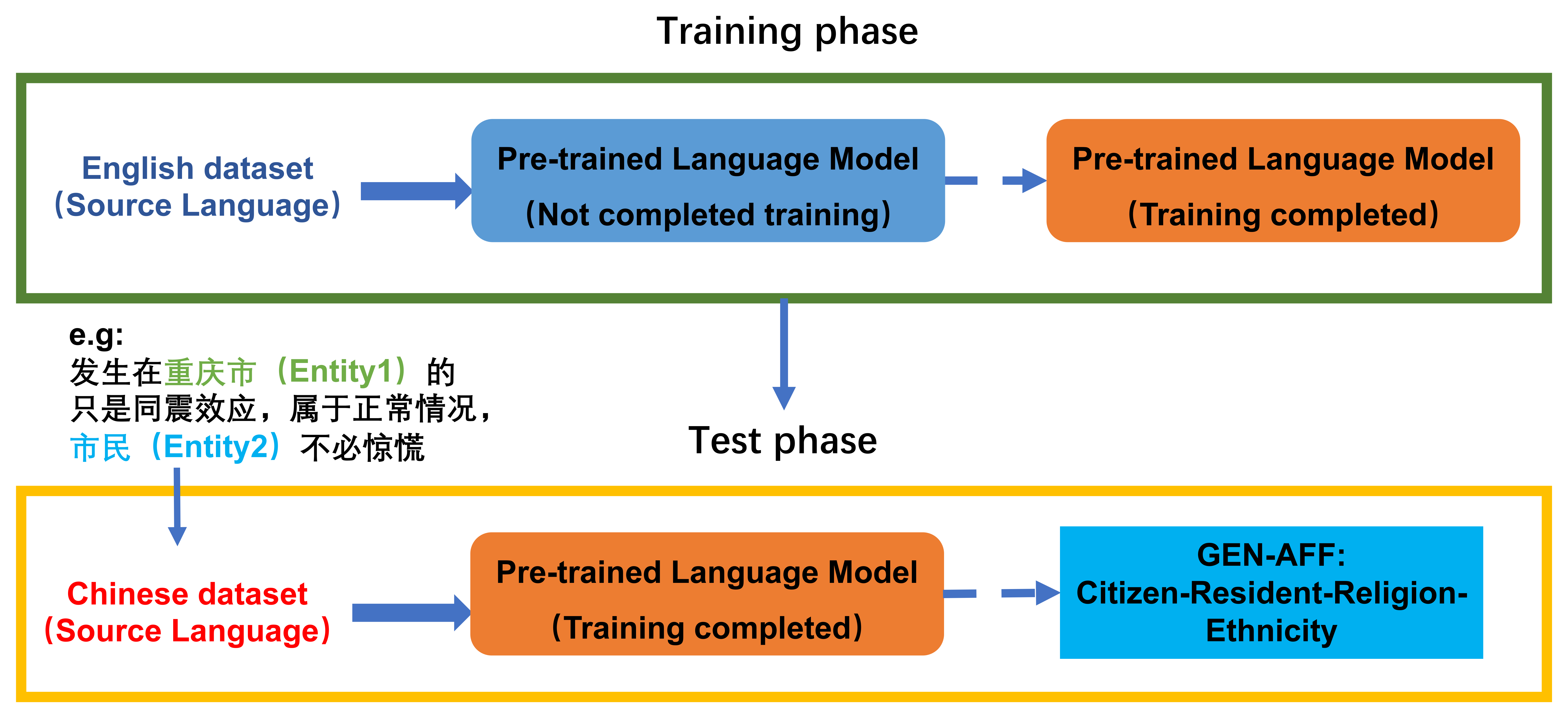}
\caption{\textbf{Zero-shot Cross-Lingual Transfer Relation Extraction.} Use English as the source language to input PLMs (Not completed training), and then Chinese as the target language to input the pre-trained Multilingual Language Model (Training completed) for testing.}  
\label{figure2}
\end{figure}
%%%%%%%%%%%%%%%%%%%%%%%%%%%%%%%%%%%%%%%%%%

\subsection{Templates}
 We choose hard-encoding templates with natural language and soft-encoding templates with additional special tokens in our work. We have designed 9 prompt-templates. The first 4. are \textbf{soft prompts}, and their input uses [MASK] to split the sentence and the two ``entities''. \textbf{Prompt\_5} and \textbf{Prompt\_6} are \textbf{hard prompt}. It is very difficult to design \textbf{hard prompt} templates for relational extraction, so we use meaningful connectives to design \textbf{hard prompt} templates that only emphasize the input ``entities'' and sentence, which we hope will help stimulate model learning and prediction. \textbf{Prompt\_7-Prompt\_9} is \textbf{hard-soft prompt}, where we combine \textbf{hard prompt} and \textbf{soft prompt}, adding meaningful conjunctions and prepositions to them and using [MASK] to split the different entities to emphasize the meaning of the entity itself.
 Normally ``[MASK]'' is used as the predicted token in prompt-tuning, but in our prompt-tuning ``[MASK]'' is a special token that is inserted to split between ``Ent1'', ``Ent2'' or ``Sent''.  The specific prompt-templates are shown in Table \ref{prompttemplates}.

 %%%%%%%%%%%%%%%%%%%%%%%%%%%%%%%%%%%%%%%%%%%%%%%%%%%%%%
\begin{table*}[]
\setlength{\tabcolsep}{4mm}{}
\centering
\begin{tabular}{ll}
\toprule
\multicolumn{2}{c}{\textbf{Designed Sequence-to-Sequence Prompts}}\\
\hline
\multicolumn{2}{c}{\textbf{\textit{Soft Prompts}}}\\
\hline
\multicolumn{1}{l}{{Prompt \_1}} & {\begin{tabular}[c]{@{}l@{}}ENC:\textless{}s\textgreater Sent {[}MASK{]}  Ent1 {[}MASK{]} Ent2 \textless{}/s\textgreater\\ DEC:\textless{}s\textgreater Sent \textless{}/s\textgreater{}\end{tabular}}\\ \hline
\multicolumn{1}{l}{{Prompt \_2}} & {\begin{tabular}[c]{@{}l@{}}ENC:\textless{}s\textgreater Sent {[}MASK{]} Ent1 {[}MASK{]} Ent2 \textless{}/s\textgreater\\ DEC:\textless{}s\textgreater Ent1 Ent2 \textless{}/s\textgreater{}\end{tabular}}\\ \hline
\multicolumn{1}{l}{{Prompt \_3}} & {\begin{tabular}[c]{@{}l@{}}ENC:\textless{}s\textgreater Sent {[}MASK{]} Ent1 {[}MASK{]} Ent2 \textless{}/s\textgreater\\ DEC:\textless{}s\textgreater Ent1 {[}MASK{]} Ent2 \textless{}/s\textgreater{}\end{tabular}}\\ \hline
\multicolumn{1}{l}{{Prompt \_4}} & {\begin{tabular}[c]{@{}l@{}}ENC:\textless{}s\textgreater Sent {[}MASK{]} Ent1 {[}MASK{]} Ent2 \textless{}/s\textgreater\\ DEC:\textless{}s\textgreater Sent {[}MASK{]} Ent1 {[}MASK{]} Ent2 \textless{}/s\textgreater{}\end{tabular}}\\ \hline
\multicolumn{2}{c}{\textbf{\textit{Hard Prompts}}}\\
\hline
\multicolumn{1}{l}{Prompt \_5} & {\begin{tabular}[c]{@{}l@{}}ENC:\textless{}s\textgreater Sent \textless{}/s\textgreater\\ DEC:\textless{}s\textgreater What is the type of relationship between Ent1 and Ent2  \textless{}/s\textgreater{}\end{tabular}}\\
\hline
\multicolumn{1}{l}{Prompt \_6} & {\begin{tabular}[c]{@{}l@{}}ENC:\textless{}s\textgreater The sentence of Sent includes Ent1 and Ent2 \textless{}/s\textgreater\\ DEC:\textless{}s\textgreater What is the type of relationship between Ent1 and Ent2  \textless{}/s\textgreater{}\end{tabular}}\\ \hline
\multicolumn{2}{c}{\textbf{\textit{Hard-Soft Prompts Hybrid Tuning}}}\\
\hline
\multicolumn{1}{l}{Prompt \_7} & {\begin{tabular}[c]{@{}l@{}}ENC:\textless{}s\textgreater The sentence of Sent includes Ent1 {[}MASK{]} Ent2 \textless{}/s\textgreater\\ DEC:\textless{}s\textgreater The sentence of Sent includes Ent1 {[}MASK{]} Ent2 \textless{}/s\textgreater{}\end{tabular}} \\ \hline
\multicolumn{1}{l}{Prompt \_8} & {\begin{tabular}[c]{@{}l@{}}ENC:\textless{}s\textgreater The sentence: ``Sent'' includes Ent1 {[}MASK{]} Ent2 \textless{}/s\textgreater\\ DEC:\textless{}s\textgreater Sent \textless{}/s\textgreater{}\end{tabular}}\\ \hline
\multicolumn{1}{l}{Prompt \_9} & {\begin{tabular}[c]{@{}l@{}}ENC:\textless{}s\textgreater The sentence: ``Sent'' includes Ent1 {[}MASK{]} Ent2 \textless{}/s\textgreater\\ DEC:\textless{}s\textgreater Ent1 {[}MASK{]} Ent2 \textless{}/s\textgreater{}\end{tabular}}\\ \bottomrule
\end{tabular}
\vspace{2mm}
\caption{\textbf{Templates for prompt-based fine-tuning on XRE task.} We design 9 different templates for our prompt-based fine-tuning algorithm, which can divide into \textbf{hard-prompts}, \textbf{soft-prompts} and \textbf{hard-soft prompts}. ``Sent'' indicts the input sentence, ``</s>'' represents the special symbol for mBART input, ``ENC'' denotes the input for encoder while ``DEC'' denotes the input for decoder.
} 
\label{prompttemplates}
\end{table*}
%%%%%%%%%%%%%%%%%%%%%%%%%%%%%%%%%%%%%%%%%%%%%%%%%%%%%%

\section{Chinese-English Relation Extraction Dataset}
\label{sec:wmtenzh}
The \textbf{WMT17} En-Zh parallel corpus~\cite{hassan2018achieving} is a high-quality high-resource dataset in machine translation community~\cite{ding2020self,ding2020context,ding2021understanding,ding-etal-2021-improving,Ding2022RedistributingLW}, which is thus considered as a preliminary dataset for RE. And since it is not annotated with relations, we took part of the dataset out of it for processing. 
To prepare the dataset for evaluation, we first parse the English sentences with the popular tool spaCy\footnote{\url{https://spacy.io/}}, and generate the connection between entities with PrePatt~\cite{white2016universal}. After Lemmatization of the extracted connections, we select the top 106 frequent links, and result in 0.9M pairs of Chinese sentences and English relations with entities. As shown in Table \ref{tab:accents2}, we randomly divided for train/validation/test set with 888k/21.2k/31.8k samples.

\section{Experiments}
\label{sec:exp}
In this section, we analyze and compare the performance of out prompt-tuning-based zero-shot XRE models on the commonly used benchmark \textbf{ACE05}~\cite{2006ACE}.

\subsection{Datasets}
\paragraph{\bf ACE05}  The experiments were conducted using the Automatic Content Extraction (ACE)2005 corpus, which includes manual annotations of relations and their parameters for three languages: English (En), Chinese (Zh), and Arabic (Ar). The ontology defined by ACE05 includes 7 entity types and 18 relation subtypes. The dataset split used in this study is the same as that in ~\cite{subburathinam2019cross}, and the preprocessing steps can be referred to in the same publication.

From the ACE05 dataset, 2607 Arabic sentences, 7124 Chinese sentences, and 6653 English sentences were extracted for this experiment, with two entities separated from each sentence. The zero-shot cross-lingual transfer approach was employed in this study, where the encoder input and the decoder input were set to the same language and the model was trained using a single language. The model was pre-trained using the ACE05 dataset and then applied to different languages for testing. The number of samples in the training/development/test set of ACE05 can be seen in Table \ref{tab:accents2}.
%%%%%%%%%%%%%%%%%%%%%%%%%%%%%%%%
\begin{table}[]
\centering
\begin{tabular}{lccc}
\toprule
\multicolumn{1}{l}{\textbf{\textit{ACE05}}} & \textbf{Train} & \textbf{Dev} & \textbf{Test} \\ \hline
\textbf{English} & 6653 & 822 & 849           \\ 
\textbf{Chinese}   & 7124 & 911 & 779           \\ 
\textbf{Arabic}  & 2067 & 304 & 305           \\ \midrule
\multicolumn{1}{l}{\textbf{\textit{WMT17-EnZh XRE}}} & \textbf{Train} & \textbf{Dev} & \textbf{Test} \\ \hline
\textbf{English-Chinese} & 888K & 21.2k & 31.8k \\
\bottomrule
\end{tabular}
% \vspace{6mm}
\caption{ \textbf{The number of documents in the \textbf{ACE05} and \textbf{WMT17-EnZh XRE} train/dev/test sets.}}
\label{tab:accents2}
\end{table}
%%%%%%%%%%%%%%%%%%%%%%%%%%%%%%%

\subsection{Experimental Setup}

\paragraph{\bf Baseline Models}
We picked certain baseline methods to compare our models on relational role categorization tasks. The baseline code for some of these methods is not publicly available, but fortunately, the work in~\cite{ahmad2021gate} this paper reproduces all of this unpublished baseline code, and here we will use their reproduce structure as a baseline.

\smallskip\noindent
{$\bullet$ CL-Trans-GCN}~\cite{liu2019neural}, each word in a phrase written in the source language is mapped to its most appropriate translation in the target language using the context-dependent lexical mapping approach known as CL-Trans-GCN~\cite{liu2019neural}. 
Along with feature embeddings that are independent of language, such as entity type embeddings, dependency relation labels, and part-of-speech (POS) tags, they also employ multilingual word embeddings to represent tokens continuously.
A baseline training is carried out for each combination of the source and target languages because this model concentrates on the target language. They directly evaluate on the target languages after training this baseline in the source languages.

\smallskip\noindent
{$\bullet$ CL-GCN}~\cite{subburathinam2019cross}
uses GCN to learn structured common space representation. We employ language-universal feature embeddings and multilingual contextual representations~\cite{devlin-etal-2019-bert} to include the tokens into an input phrase.

\smallskip\noindent
{$\bullet$ Bi-LSTM} the Bi-Directional Long Short-Term Memory (Bi-LSTM) based RE model.

\smallskip\noindent
{$\bullet$ CL-RNN}~\cite{florian2019neural} uses a Bi-LSTM type recurrnet neural networks to learn contextual representation.
They provide words in a sentence that is put together similarly to~\cite{subburathinam2019cross} with language-universal features. Similar to CL-GCN, they train and assess this baseline.

\smallskip\noindent
{$\bullet$ Transformer}~\cite{vaswani2017attention} uses multi-head self-attention mechanism.

\smallskip\noindent
{$\bullet$ Transformer-RPR }~\cite{shaw-etal-2018-self} encodes the structure of the input sequences using relative position representations. It employs pairwise sequential distances.

\smallskip\noindent
{$\bullet$ GATE}~\cite{ahmad2021gate}, Graph Attention Transformer Encoder, and test its cross-lingual transferability on RE.

\smallskip\noindent
{$\bullet$ CNN}, the Convolutional Neural Network (CNN) based XRE model.

\smallskip\noindent
{$\bullet$ XLM-R}, RE models based on XLM-R that use Sentence Start (SS), Entity Start (ES), Entity Max Pooling (EMP), and Uniform Markers (UM) or Entity Type Markers (ETM) as well as summary representation.

\smallskip\noindent
{$\bullet$ mBERT} based RE models that use Entity Start (ES), Entity Max Pooling (EMP), and Sentence Start (SS) as summary representations~\cite{2019Matching}.

\subsection{Main Results}
\paragraph{\bf Monolingual RE Performance} Before doing  XRE, we first performed monolingual relation extraction using the model. For monolingual training, we utilize the mBART and mBART+prompt models, and for testing, we choose the top model from the validation set loss. The results are displayed in Table\ref{table:Monolingual-re-results}, the performance of our mBART+prompt model is noteworthy, especially when the source and target languages are the same. As demonstrated in the results, our models outperformed the first three baseline models by $+12.2$, $+8.8$, and $+16.6$ when the language was English. Similarly, when the language was Arabic, our models outperformed the first three models by $+9.4$, $+7.4$, and $+12.0$. Furthermore, our models demonstrated superior performance over the first three models by $+16.3$, $+13.0$, and $+3.6$ when the language was Chinese.
In terms of the mean, our model outperformed the three baseline models, achieving $+12.6$, $+9.7$, and $+10.8$ more points, respectively. These results demonstrate the efficacy of our approach, which combines the mBART and prompt models to achieve improved performance in monolingual relation extraction.

%%%%%%%%%%%%%%%%%%%%%%%%%%%%%%%%%%%%%%%%%%%%%%%%%%%%%%%%%%%%%%%%
\begin{table*}
\small
\begin{center}
\begin{tabular}{l|cccc}
\toprule \textbf{Model} & \multicolumn {4}{c}{\textbf{ACE05}}\\
\cline{2-5} & \textbf{En} & \textbf{Ar} & \textbf{Zh} & {\it Avg.}  \\
\midrule
mBERT-ETM-EMP~\cite{2019Matching} (Monolingual) & 70.3 & 70.9 & 71.7 & \underline{71.0}  \\
XLM-R-ETM-EMP (Monolingual) &  73.7 & 72.9 & 75.0 & \underline{73.9} \\
mBART (Monolingual) &  65.9 & 68.3 & 84.4 & \underline{72.8} \\
mBART+prompt (Monolingual) &  \textbf{82.5} & \textbf{80.3} & \textbf{88.0} & \underline{\textbf{83.6}} \\
\bottomrule
\end{tabular}
\vspace{2mm}
\end{center}
\vspace*{-1mm}
\caption{\textbf{Results of RE models in monolingual (a unique model is trained using data from each language).} Average scores on all delays are \underline{underlined}. The best result is \textbf{bold}.} \label{table:Monolingual-re-results}
\vspace*{-3mm}
\end{table*}
%%%%%%%%%%%%%%%%%%%%%%%%%%%%%%%%%%%%%%%%%%%%%%%%%%%%%%%%%%%%%%%

\paragraph{\bf XRE Performance} Each of the three languages RE models are used with additional target languages. For cross-lingual testing of monolingual-based language models, we choose the best model from the validation set loss. Because they were trained in the same embedding space, these models may be utilized instantly to conduct zero-shot XRE on additional languages. (the cross-lingual representation learning framework).

As can be seen in Table \ref{table:gate_vs_trans}, our XRE models outperform models in earlier works in terms of accuracy, achieving state-of-the-art XRE performance across all languages. The missing values in Table \ref{table:gate_vs_trans} are due to the absence of experimental data for those models, which cannot be displayed.
In particular, the mean value of our model f1-score is 68.7 in all 3 target languages, which is much higher than the rest of the baseline models. The f1-scores of \textbf{En-Zh}, \textbf{En-Ar}, \textbf{Ar-Zh} and \textbf{Zh-Ar} reached 77.2, 67.5, 69.4, and 63.6, respectively, which are $+22.1$, $+0.7$, $+15.1$ and, $+2.4$ higher than \textbf{GATE}, respectively. It is interesting to note that when the target language is Chinese, our XRE model improves very much compared to the rest of the baseline models, which is very obvious when the model uses English as the pre-training language. Our XRE models obtain significant gains compared with other baseline models, especially for the model pre-trained with English.
It is also found from the table that when the target language is Arabic, our XRE model also shows a large improvement.

The performance of the approaches in the table for the two cases \textbf{En-Zh} and \textbf{En-Ar} is then compared. There are some baseline methods in the table that have only \textbf{En-Zh} and \textbf{En-Ar} results, this is because these baseline methods have only the above two results. Our model still performs very well for both \textbf{En-Zh} and \textbf{En-Ar} XRE tasks. When Arabic is the target language, our model performs significantly better than previous models.

Adding prompt templates can improve the cross-lingual relation extraction capability of the \textbf{mBART}. This is because prompt templates introduce language-specific information into the model, allowing it to better understand relationships between different languages. By using prompt templates, the model can learn how to map relationships across different languages, leading to better cross-lingual relation extraction.

Furthermore, prompt templates can help mitigate the effects of data sparsity, which is particularly important in cross-lingual relation extraction. Due to imbalanced training data across different languages, using prompt templates can help the model better utilize limited data and perform more accurate cross-lingual relation extraction.

%%%%%%%%%%%%%%%%%%%%%%%%%%%%%%%%%%%%%%%%%%%%%%%%%%%%%%%%%%%%%%%%
\begin{table*}[t]
\centering
% \resizebox{0.95\linewidth}{!}{%
\scriptsize
\setlength{\tabcolsep}{4mm}{} 
\begin{tabular}{l|cccccc|c}
\toprule
\multirow{1}{*}{\bf Model}
% \multicolumn{6}{c}{Relation Extraction}
% \\ \cline{5-10}
& \bf En-Zh & \bf En-Ar & \bf Zh-En & \bf Zh-Ar & \bf Ar-En & \bf Ar-Zh & \it Avg. \\ 
% & $\Downarrow$ & $\Downarrow$ & $\Downarrow$ & $\Downarrow$ & $\Downarrow$ & $\Downarrow$ & $\Downarrow$ & $\Downarrow$ & $\Downarrow$ \\
% & Ar & Zh & En & Zh & Ar & En & Ar & En & Zh \\ 
\midrule
{\textbf{Transformer}} & 57.1 & 63.4 & 69.6 & 60.6 & 67.0 & 52.6 & \underline{61.7}\\
{\textbf{Transformer RPR}} & 58.0 & 59.9 & 70.0 & 55.6 & 66.5 & 56.5 & \underline{61.1}\\
{\textbf{GATE}} & {55.1} & 66.8 & \textbf{71.5} & 61.2 & \textbf{69.0} & 54.3 & \underline{63.0}\\
{\textbf{CL Trans GCN}} & 56.7 & 65.3 & 65.9 & 59.7 & 59.6 & 46.3 & \underline{58.9}\\
{\textbf{CL GCN}} & 49.4 & 58.3 & 65.0 & 55.0 & 56.7 & 42.4 & \underline{54.5}\\
{\textbf{CL RNN}} & 53.7 & 63.9 & 70.9 & 57.6 & 67.1 & 55.7 & \underline{61.5}\\
{\textbf{mBERT-UM-ES}} & 58.7 & 38.1 & - & - & - & - & -\\
{\textbf{mBERT-UM-EMP}} & 59.7 & 38.2 & - & - & - & - & -\\
{\textbf{XLM-R-UM-ES}} & 62.9 & 46.8 & - & - & - & - & -\\
{\textbf{XLM-R-UM-EMP}} & 63.5 & 44.9 & - & - & - & - & -\\
{\textbf{XLM-R-ETM-ES}} & 65.7 & 46.2 & - & - & - & - & -\\
{\textbf{XLM-R-ETM-EMP}} & 64.9 & 49.7 & - & - & - & - & -\\
{\textbf{mBART}} & 57.4 & 51.3 & 69.1 & 53.4 & 68.1 & 57.8 & \underline{59.5}\\
{\textbf{mBART+prompt}} & \textbf{77.2} & \textbf{67.5} & 69.4 & \textbf{63.6} & 65.9 & \textbf{67.3} & \underline{\textbf{69.0}}  \\
\bottomrule
\end{tabular}
% }
\vspace{2mm}
\caption{
\textbf{Results (F-score \% on the test set) of various models tested on the \textbf{ACE05} dataset.}
The language on left and right of $\Rightarrow$ denotes the source and target languages, respectively. Average scores on all delays are \underline{underlined}. The best result is \textbf{bold}.
}
\label{table:gate_vs_trans}
\vspace{-2mm}
\end{table*}

%%%%%%%%%%%%%%%%%%%%%%%%%%%%%%%%%%%%%%%%%%%%%%%
\section{Analysis}
\label{sec:analysis}
In this section, we conduct an ablation study to test that our prompt template worked reasonably well in experiments based on the mBART model. We also compare the effects of \textbf{soft prompt}, \textbf{hard prompt} and, \textbf{hard-soft prompt} on the experimental results, conduct experiments on the \textbf{WMT17-EnZh XRE} dataset and analyze the experimental results to demonstrate the experimental capability of our constructed dataset and the adaptability of our model and prompt template to large datasets.

\subsection{Ablation Study}
As shown in Table \ref{prompt-re-result}, We can see that all the F1-scores from our model are higher than the f1-scores from running the mBART model alone. We can see from the average that the XRE model with the 9 prompts all perform better than the mBART model alone.
In the Ar-Ar, except for Prompt\_5, all the prompts performed higher than mBART, especially the F1-score of \textbf{Prompt\_7} was higher than mBART $+12$ points.
In the En-En, all prompts outperform mBART, with the F1-score of \textbf{Prompt\_4} and \textbf{Prompt\_8} both being 82.5 and F1-score $+16.6$ points higher than mBART.
In the Zh-Zh, all the prompts outperform mBART, with \textbf{Prompt\_6} having the best performance and F1-score $+3.6$ points higher than mBART.
In the Ar-En, Prompt\_2, 3, 7, 8 outperform mBART, with \textbf{Prompt\_3} performing the best and its F1-score $+3.2$ points higher than mBART.
In Ar-Zh, Prompt\_2, 3, 4, 5, 7, 8 perform better than mBART, especially \textbf{Prompt\_2} perform the best, its F1-score was 67.3 points higher than mBART $+9.5$ points.
In En-Ar, all prompt experiments perform better than mBART, with \textbf{Prompt\_7} performing the best f1 value of 67.5, $+16.2$ points higher than the f1 score of the mBART model.
In En-Zh,  all prompts perform better than mBART, especially \textbf{Prompt\_7} performs best, its F1-score is 77.2 points higher than mBART $+19.8$ points.
In Zh-Ar, except for Prompt\_5, Prompt\_6, Prompt\_7, all of them perform better than mBART, especially \textbf{Prompt\_9} has the best performance, its F1-score is 63.6 points higher than mBART $+10.2$ points
In Zh-En, Only \textbf{Prompt\_4} is higher than mBART $+0.3$ points. Comparing the average F1-score of all prompt, Prompt\_2, 3, 4, 7, 9 all have an average score of 70 or more. Especially \textbf{Prompt\_3} average reach a maximum of 70.9 points higher than mBART $+12.3$ points

The above experimental results show that our prompt template is effective for improving the model monolingual or XRE. This also proves that our prompt templates are huge improvement on the original mBART for this task.

%%%%%%%%%%%%%%%%%%%%%%%%%%%%%%%%%%%%%%%%%%%%%%%%%%%%%%%%%%%%%%%%
\begin{table*}[t]
\begin{center}
\footnotesize
\setlength{\tabcolsep}{4mm}{}
\centering
\begin{tabular}{lcccccccccc}
\toprule
% \multicolumn{11}{|c|}{\textbf{ACE05}}                           \\ \midrule
{\textbf{Model}}&
{\textbf{Ar-Ar}} &
{\textbf{En-En}} &
{\textbf{Zh-Zh}} &
{\textbf{Ar-En}} & 
{\textbf{Ar-Zh}} & 
{\textbf{En-Ar}} & 
{\textbf{En-Zh}} & 
{\textbf{Zh-Ar}} & 
{\textbf{Zh-En}} &  
\textbf{Avg.}\\ \midrule
{\textbf{mBART}}& 
{68.3} & 
{65.9} &
{84.4} &
{62.7} & 
{57.8} &
{51.3} & 
{57.4} &
{53.4} & 
{69.1} &   
\underline{58.6}  \\
\hline
\multicolumn{11}{c}{\textbf{\textit{Soft Prompts}}}\\
\hline
{\textbf{Prompt\_1}}& 
{72.7} & 
{80.9} &
{86.7} &
{61.4} & 
{53.6} &
{62.9} & 
{77.1} &
{55.7} & 
{67.4} &   
\underline{68.7}  \\
{\textbf{Prompt\_2}}& 
{78.0} & 
{82.3} &
{86.0} &
{65.0} & 
{\textbf{67.3}}     & 
{58.0} & 
{74.7} & 
{58.0} & 
{68.5} &   
\underline{70.8}  \\ 
{\textbf{Prompt\_3}} & 
{75.7} & 
{82.3} &
{86.0} &
{\textbf{65.9}} & 
{65.8} & 
{64.2} & 
{74.4} & 
{58.3} & 
{65.8} &    
\underline{\textbf{70.9}}   \\ 
{\textbf{Prompt\_4}} &
{78.6} & 
{\textbf{82.5}} &
{85.1} & 
{60.3} & 
{62.3} & 
{64.5} & 
{73.8} & 
{58.0} & 
{\textbf{69.4}} &    
\underline{70.5}     \\
\hline
\multicolumn{11}{c}{\textbf{\textit{Hard prompts}}}\\
\hline
{\textbf{Prompt\_5}} & 
{62.4} & 
{79.6} &
{80.2} &
{54.6} & 
{59.3} & 
{60.4} & 
{70.5} & 
{53.5} & 
{62.1} &    
\underline{64.7}   \\
{\textbf{Prompt\_6}} &
{79.3} &
{81.1} &
\textbf{88.0} &
{57.0} &
{57.4} &
{63.0} &
{77.0} &
{53.1} &
{67.2} &
\underline{69.2} \\
\hline
\multicolumn{11}{c}{\textbf{\textit{Hard-Soft Prompt}}}\\
\hline
{\textbf{Prompt\_7}} & 
{\textbf{80.3}} & {81.2} &
{87.4} &
{65.2}           & {59.3}           & {\textbf{67.5}}           & {76.6}              & {49.1}           & {68.4}         & \underline{70.5}  \\
{\textbf{Prompt\_8}} &
{74.4} & {\textbf{82.5}} &
{85.8} &
{64.0} & {59.1} & {60.3} & {76.1} & {54.4} & {66.7} &    \underline{69.2}    \\
{\textbf{Prompt\_9}} &
{77.3} & 
{81.9} & 
{86.5} &
{60.5} & 
{57.2} & 
{60.6} & 
{\textbf{77.2}} & 
{\textbf{63.6}} & 
{69.0} &   
\underline{70.4}  \\ 

\bottomrule
\end{tabular}
\vspace{2mm}
\end{center}
\caption{\textbf{Performance ($F_1$ score) of XRE models on the \textbf{ACE05} dataset.} Average scores on all delays are \underline{underlined}. The best result is \textbf{bold}.}\label{prompt-re-result}
\end{table*}
%%%%%%%%%%%%%%%%%%%%%%%%%%%%%%%%%%%%%%%%%%%%%%%%%%%%%%%%%%%%%%%%%%%

\subsection{Effects on Large Dataset} As the data scale \textbf{ACE05} is limited, it is natural to doubt the effectiveness of our prompt learning strategy on the large-scale dataset. On the other hand, as the lack of XRE datasets based on parallel corpora, we construct this dataset to validate its effect on this dataset. To this end, we conduct experiments on our self-construct \textbf{WMT En-Zh} (see Section~\ref{sec:wmtenzh}) XRE dataset. We randomly divid for train/validation/test set with 888k/21.2k/31.8k samples.

\paragraph{\bf WMT RE} In the \textbf{WMT RE} dataset we choose Chinese as the source language and English as the target language. Since this dataset is a parallel corpus, we do not use the zero-shot setting for this dataset in the pre-training phase. We set the encoder input to Chinese and the decoder input to English for the experiment. As shown in Table \ref{prompt-re-result}, \textbf{soft prompt} and \textbf{hard-soft prompt} perform well, so we did not use \textbf{hard prompt} for our experiments.

The results of our XRE model with mBART on the WMT RE dataset are shown in Table \ref{wmt-re}. We selected several prompt templates for evaluation, among which, \textbf{Prompt\_4} and \textbf{Prompt\_8} showed superior performance. These results demonstrate the generalizability and adaptability of our XRE model to datasets of varying sizes and types.

\subsection{The Effects of Different prompt}
This section compares how different prompt kinds affected the outcomes of the \textbf{ACE05} experiment.

First, we compare the effect of different \textbf{soft prompts} on the experimental results. The encoder input is the same for all \textbf{soft prompts}, but the difference lies in the input of the decoder. Prompt\_1 and Prompt\_2 are compared from the experimental results, Prompt\_1 is better than Prompt\_2 only in the two tasks of En-Ar and En-Zh, comparing their decoder inputs, Prompt\_1 is a sentence and Prompt\_2 is two ``entities''. This may be because there are many entities in the same sentence, and these entities have different relationships, so Prompt\_2 with two entities in the decoder input is more responsive to the relationship of entities in the sentence. 

In Prompt\_3, the input to the decoder is two ``entities'' and ``[MASK]'' tokens, while Prompt\_4 adds a sentence. The purpose of the ``[MASK]'' token is to distinguish the ``entity'' from the sentence and help the model identify them for better prediction. Although we have a significant improvement over mBART, we expect that the addition of sentences to Prompt\_3 will improve the ability to predict relationship types. Only two tasks, Zh-En, and En-Ar, have been improved compared with Prompt\_3. The performance of the above \textbf{soft prompt} compared to mBART is superior to mBART in terms of experimental results.

Next, we compare the difference in the performance of different \textbf{hard prompt} in the RE task. The difference between Prompt\_5 and Prompt\_6 is the encoder input. The encoder input for Prompt\_5 is sentence and the encoder input for Prompt\_6 is a meaningful word, the preposition sentence, and two ``entities'' in the sentence. The encoder input of Prompt\_6 emphasizes the entities in the sentence, while the encoder input of Prompt\_5 only has a sentence and does not emphasize the entities, so the performance of Prompt\_6 is better than Prompt\_5 from the experimental results. This demonstrates that ``entity'' is crucial to the understanding of entity relations. We can also see from the experiments that both \textbf{hard prompts} we have designed perform better than mBART on average for the RE task in different languages.

Finally, we compare the effectiveness of several different \textbf{hard-soft prompt} for the extraction task.
In Prompt\_7, Prompt\_8, and Prompt\_9, we added English cue words and ``[MASK]'' tokens to the encoder input. We hope and believe that the cue stimulates the ability of the model to understand and predict the sentence, while the [MASK] token is used to segment the sentence and ``entities'' in the sentence for the model to distinguish them and improve the model's prediction ability. The encoder of Prompt\_7 has the same input as the decoder and emphasizes entities. According to the findings of the investigation, Prompt\_7 and Prompt\_9 perform better than Prompt\_8, while Prompt\_8 performs worse.  This may be because the input of Prompt\_8 only has a sentence, and since there are many ``entities'' in a sentence, the relationship types between different entities are different, so it causes the model to fail to accurately predict the relationship types of entities in the model when predicting.

The experimental results and subsequent analysis indicate that both \textbf{soft prompt} and \textbf{hard-soft prompt} outperform \textbf{hard prompt}, with \textbf{soft prompt} particularly exhibiting strong performance. The versatility of \textbf{soft prompt} templates, which are not limited by natural language, contributes to their superiority over \textbf{hard prompt} and \textbf{hard-soft prompt}, and makes them a suitable choice for cross-lingual relation extraction tasks.

\begin{table}
\centering
\begin{tabular}{lcc}
\toprule
% \multicolumn{2}{|c|}{\textbf{WMT17}}                           \\ \hline
{\textbf{Model}}  &  \textbf{Acc.} & \textbf{$\Delta$}\\ \midrule
% {\textbf{lstm}} & 40.60 \\
\textbf{mBART} & 47.1 & -\\
{\textbf{Prompt\_1}} &  64.6 & +17.5 \\
{\textbf{Prompt\_3}} &  64.8 & +17.7 \\ 
{\textbf{Prompt\_4}} &  \textbf{65.4} & +18.3 \\
{\textbf{Prompt\_7}} &  64.8 & +17.7 \\
{\textbf{Prompt\_8}} &  65.2  & +18.1 \\
{\textbf{Prompt\_9}} &  64.5 & +17.4 \\
\bottomrule
\end{tabular}
\vspace{2mm}
\caption{\textbf{Performance (accuracy) of XRE models on the large scale WMT RE dataset.} The best result is \textbf{bold}.}\label{prompt-wmt17-result}
\label{wmt-re}
\end{table}

\section{Relate work}
\label{sec:relate}

\subsection{Fine-tuning with Pre-trained Multilingual Language Model}
Various recent PLMs~\cite{vega1,vega2} like BERT and GPT provide a new paradigm for utilizing large-scale unlabeled data for NLP tasks. Although these PLMs can capture rich knowledge~\cite{yenicelik-etal-2020-bert}  from massive corpora, a fine-tuning process with extra task-specific data is still required to transfer their knowledge for downstream tasks. From dialogue~\cite{wu2020slotrefine,wu2021bridging, cao2021towards}, summarization~\cite{zan2022bridging,zhang2022bliss}, question answering~\cite{zhong2022improving,zhong2022e2s2}, retrieval~\cite{rao2022does,rao2023dynamic}, text classification~\cite{Wang2022ACC,zhong2023knowledge} to machine translation~\cite{vegamt,zan2022complementarity}, fine-tuned PLMs have been demonstrated their effectiveness on almost all important NLP tasks. Besides fine-tuning language models for specific tasks, recent studies have explored better optimization and regularization techniques to improve fine-tuning~\cite{2020Fine}.

\subsection{Prompt-Learning}
A brand-new paradigm in PLMs~\cite{liu2021pre} called prompt-based learning is currently prevailing and effective.
Prompt-based methods~\cite{zhong2022panda}, in contrast to the pre-training and fine-tuning paradigm, convert downstream tasks to a form more consistent with the model's pre-training tasks.
\cite{schick2020exploiting} convert a variety of classification problems to cloze tasks by building related prompts with blanks and determining a mapping from specific filled words to predict categories.
By freezing model parameters and just modifying a series of continuous task-specific vectors,~\cite{li2021prefix} proposes lightweight prefix tweaking with a focus on generation tasks.
In natural language processing, prompting-based techniques have shown promise as a novel paradigm for zero-shot or few-shot inference~\cite{liu2021pre}.
Another benefit is the capability to easily adapt very large language models~\cite{reynolds2021prompt} to new tasks without the need for comparatively expensive fine-tuning.

\subsection{Zero-Shot Relation Extraction} 
Previously, zero-shot relation extraction was approached as a slot-filling task~\cite{levy2017zero} and solved through the use of reading comprehension techniques. However, because each relation label in their method must be manually designed, it cannot adapt well to new relation types. The formulation of zero-shot RE into an entailment task~\cite{obamuyide-vlachos-2018-zero}, which is not restricted to a fixed relation label space, is another method. The entailment approach instead determines whether or not any two sentences or relation labels are compatible.
Data augmentation is another well-liked method for enhancing model performance in supervised lower-source tasks. The quality of augmented samples was initially improved by simple heuristics like token manipulation, and new approaches to language modeling further enhanced the quality.
While there are data augmentation techniques that can be used for structured tasks like named entity recognition and RE, they require pre-existing training samples and are difficult to adapt to zero-shot tasks.

\section{Conclusion}

In this work, we introduce a novel approach to zero-shot cross-lingual relation extraction (XRE) using prompt learning techniques upon multilingual pretrained language models. To design the prompt templates, we carefully propose three different methods: \textbf{hard prompt}, \textbf{soft prompt}, and \textbf{hard-soft prompt}. Among the prompts we designed, the soft prompt -- \textbf{Prompt\_3} emphasizes practicality and generalization, and avoids limiting the model to a single language. 
The results of our experiments validate the effectiveness of our proposed model, and we also conducted extensive tests to thoroughly analyze where the performance gain comes from.

Due to the limited availability of parallel corpus-based XRE datasets, we created a new RE dataset by extracting English-Chinese dataset from the \textbf{WMT17 En-Zh} parallel corpus, resulting in a dataset of size 0.9M. Our experiments on large datasets demonstrate the efficacy and generalizability of our prompt-learning strategy.
In the future, we plan to further investigate the design of prompt templates for different languages, as well as the combination of pre-trained multilingual language models with prompt-tuning.

\section{Acknowledgements}
This work was supported in part by the Natural Science Foundation of China (Nos. 61976162, 82174230), Artificial Intelligence Innovation Project of Wuhan Science and Technology Bureau (No.2022010702040070), Science and Technology Major Project of Hubei Province (Next Generation AI Technologies) (No. 2019AEA170).
\bibliographystyle{IEEEtran}
\bibliography{IEEEabrv,reference}

\begin{thebibliography}{10}
\providecommand{\url}[1]{#1}
\csname url@samestyle\endcsname
\providecommand{\newblock}{\relax}
\providecommand{\bibinfo}[2]{#2}
\providecommand{\BIBentrySTDinterwordspacing}{\spaceskip=0pt\relax}
\providecommand{\BIBentryALTinterwordstretchfactor}{4}
\providecommand{\BIBentryALTinterwordspacing}{\spaceskip=\fontdimen2\font plus
\BIBentryALTinterwordstretchfactor\fontdimen3\font minus
  \fontdimen4\font\relax}
\providecommand{\BIBforeignlanguage}[2]{{%
\expandafter\ifx\csname l@#1\endcsname\relax
\typeout{** WARNING: IEEEtran.bst: No hyphenation pattern has been}%
\typeout{** loaded for the language `#1'. Using the pattern for}%
\typeout{** the default language instead.}%
\else
\language=\csname l@#1\endcsname
\fi
#2}}
\providecommand{\BIBdecl}{\relax}
\BIBdecl

\bibitem{2020Improving}
K.~Zhou, W.~X. Zhao, S.~Bian, Y.~Zhou, and J.~Yu, ``Improving conversational
  recommender systems via knowledge graph based semantic fusion,'' in
  \emph{KDD}, 2020.

\bibitem{Wang2022ACC}
B.~Wang, L.~Ding, Q.~Zhong, X.~Li, and D.~Tao, ``A contrastive cross-channel
  data augmentation framework for aspect-based sentiment analysis,'' in
  \emph{COLING}, 2022.

\bibitem{2021QA}
M.~Yasunaga, H.~Ren, A.~Bosselut, P.~Liang, and J.~Leskovec, ``Qa-gnn:
  Reasoning with language models and knowledge graphs for question answering,''
  \emph{arXiv preprint}, 2021.

\bibitem{ding2021progressive}
L.~Ding, L.~Wang \emph{et~al.}, ``Progressive multi-granularity training for
  non-autoregressive translation,'' in \emph{Findings of ACL}, 2021.

\bibitem{doddington-etal-2004-automatic}
G.~Doddington, A.~Mitchell \emph{et~al.}, ``The automatic content extraction
  ({ACE}) program {--} tasks, data, and evaluation,'' in \emph{LREC}, 2004.

\bibitem{10.1145/3293318}
W.~Wang, V.~W. Zheng, H.~Yu, and C.~Miao, ``A survey of zero-shot learning:
  Settings, methods, and applications,'' \emph{ACM TIST}, 2019.

\bibitem{miwa-bansal-2016-end}
M.~Miwa and M.~Bansal, ``End-to-end relation extraction using {LSTM}s on
  sequences and tree structures,'' in \emph{ACL}, 2016.

\bibitem{ahmad2021gate}
W.~U. Ahmad, N.~Peng, and K.-W. Chang, ``Gate: graph attention transformer
  encoder for cross-lingual relation and event extraction,'' in \emph{AAAI},
  2021.

\bibitem{liu2020multilingual}
Y.~Liu, J.~Gu, N.~Goyal, X.~Li, S.~Edunov, M.~Ghazvininejad, M.~Lewis, and
  L.~Zettlemoyer, ``Multilingual denoising pre-training for neural machine
  translation,'' \emph{TACL}, 2020.

\bibitem{yan2021unified}
H.~Yan, T.~Gui, J.~Dai, Q.~Guo, Z.~Zhang, and X.~Qiu, ``A unified generative
  framework for various ner subtasks,'' \emph{arXiv preprint}, 2021.

\bibitem{brown2020language}
T.~Brown, B.~Mann, N.~Ryder, M.~Subbiah, J.~D. Kaplan, P.~Dhariwal,
  A.~Neelakantan, P.~Shyam, G.~Sastry, A.~Askell \emph{et~al.}, ``Language
  models are few-shot learners,'' \emph{NeurIPS}, 2020.

\bibitem{zhang2021differentiable}
N.~Zhang, L.~Li, X.~Chen, S.~Deng, Z.~Bi, C.~Tan, F.~Huang, and H.~Chen,
  ``Differentiable prompt makes pre-trained language models better few-shot
  learners,'' \emph{arXiv preprint}, 2021.

\bibitem{2006ACE}
C.~Walker, S.~Strassel, J.~Medero, and K.~Maeda, ``Ace 2005 multilingual
  training corpus,'' \emph{Progress of Theoretical Physics Supplement}, 2006.

\bibitem{broscheit2017openie}
S.~Broscheit, K.~Gashteovski, and M.~Achenbach, ``Openie for slot filling at
  tac kbp 2017-system description.'' in \emph{TAC}, 2017.

\bibitem{zhong2023chat}
Q.~Zhong, L.~Ding \emph{et~al.}, ``Can chatgpt understand too? a comparative
  study on chatgpt and fine-tuned bert,'' \emph{arXiv preprint}, 2023.

\bibitem{Lu2023EAPrompt}
Q.~Lu, B.~Qiu, L.~Ding, L.~Xie, and D.~Tao, ``Error analysis prompting enables
  human-like translation evaluation in large language models: A case study on
  chatgpt,'' \emph{arXiv preprint}, 2023.

\bibitem{Peng2023ChatGPT4MT}
K.~Peng, L.~Ding \emph{et~al.}, ``Towards making the most of chatgpt for
  machine translation,'' \emph{arxiv preprint}, 2023.

\bibitem{hu2021knowledgeable}
S.~Hu, N.~Ding, H.~Wang, Z.~Liu, J.~Li, and M.~Sun, ``Knowledgeable
  prompt-tuning: Incorporating knowledge into prompt verbalizer for text
  classification,'' \emph{arXiv preprint}, 2021.

\bibitem{gao2020making}
T.~Gao, A.~Fisch, and D.~Chen, ``Making pre-trained language models better
  few-shot learners,'' \emph{arXiv preprint}, 2020.

\bibitem{shin2020autoprompt}
T.~Shin, Y.~Razeghi, R.~L. Logan~IV, E.~Wallace, and S.~Singh, ``Autoprompt:
  Eliciting knowledge from language models with automatically generated
  prompts,'' \emph{arXiv preprint}, 2020.

\bibitem{li2021prefix}
X.~L. Li and P.~Liang, ``Prefix-tuning: Optimizing continuous prompts for
  generation,'' \emph{arXiv preprint}, 2021.

\bibitem{tacl_mBART}
Y.~Liu, J.~Gu, N.~Goyal, X.~Li, S.~Edunov, M.~Ghazvininejad, M.~Lewis, and
  L.~Zettlemoyer, ``{Multilingual Denoising Pre-training for Neural Machine
  Translation},'' \emph{TACL}, 2020.

\bibitem{liu2021understanding}
X.~Liu, L.~Wang \emph{et~al.}, ``Understanding and improving encoder layer
  fusion in sequence-to-sequence learning,'' in \emph{ICLR}, 2021.

\bibitem{lewis2019bart}
M.~Lewis, Y.~Liu \emph{et~al.}, ``Bart: Denoising sequence-to-sequence
  pre-training for natural language generation, translation, and
  comprehension,'' \emph{arXiv preprint}, 2019.

\bibitem{hassan2018achieving}
H.~Hassan, A.~Aue \emph{et~al.}, ``Achieving human parity on automatic chinese
  to english news translation,'' \emph{arXiv preprint}, 2018.

\bibitem{ding2020self}
L.~Ding, L.~Wang, and D.~Tao, ``Self-attention with cross-lingual position
  representation,'' in \emph{ACL}, 2020.

\bibitem{ding2020context}
L.~Ding, L.~Wang, D.~Wu, D.~Tao, and Z.~Tu, ``Context-aware cross-attention for
  non-autoregressive translation,'' in \emph{COLING}, 2020.

\bibitem{ding2021understanding}
L.~Ding, L.~Wang \emph{et~al.}, ``Understanding and improving lexical choice in
  non-autoregressive translation,'' in \emph{ICLR}, 2021.

\bibitem{ding-etal-2021-improving}
L.~Ding, D.~Wu, and D.~Tao, ``Improving neural machine translation by
  bidirectional training,'' in \emph{EMNLP}, 2021.

\bibitem{Ding2022RedistributingLW}
L.~Ding \emph{et~al.}, ``Redistributing low-frequency words: Making the most of
  monolingual data in non-autoregressive translation,'' in \emph{ACL}, 2022.

\bibitem{white2016universal}
A.~S. White, D.~Reisinger, K.~Sakaguchi, T.~Vieira, S.~Zhang, R.~Rudinger,
  K.~Rawlins, and B.~Van~Durme, ``Universal decompositional semantics on
  universal dependencies,'' in \emph{EMNLP}, 2016.

\bibitem{subburathinam2019cross}
A.~Subburathinam, D.~Lu, H.~Ji \emph{et~al.}, ``Cross-lingual structure
  transfer for relation and event extraction,'' in \emph{EMNLP}, 2019.

\bibitem{liu2019neural}
J.~Liu, Y.~Chen, K.~Liu, and J.~Zhao, ``Neural cross-lingual event detection
  with minimal parallel resources,'' in \emph{EMNLP}, 2019.

\bibitem{devlin-etal-2019-bert}
J.~Devlin, M.-W. Chang, K.~Lee, and K.~Toutanova, ``{BERT}: Pre-training of
  deep bidirectional transformers for language understanding,'' in
  \emph{NAACL}, 2019.

\bibitem{florian2019neural}
J.~Ni and R.~Florian, ``Neural cross-lingual relation extraction based on
  bilingual word embedding mapping,'' in \emph{EMNLP}, 2019.

\bibitem{vaswani2017attention}
A.~Vaswani, N.~Shazeer \emph{et~al.}, ``Attention is all you need,'' in
  \emph{NeurIPS}, 2017.

\bibitem{shaw-etal-2018-self}
P.~Shaw, J.~Uszkoreit, and A.~Vaswani, ``Self-attention with relative position
  representations,'' in \emph{AACL}, 2018.

\bibitem{2019Matching}
L.~B. Soares, N.~Fitzgerald, J.~Ling, and T.~Kwiatkowski, ``Matching the
  blanks: Distributional similarity for relation learning,'' 2019.

\bibitem{vega1}
Q.~Zhong, L.~Ding \emph{et~al.}, ``Bag of tricks for effective language model
  pretraining and downstream adaptation: A case study on glue,'' \emph{arXiv
  preprint}, 2023.

\bibitem{vega2}
Q.~Zhong \emph{et~al.}, ``Toward efficient language model pretraining and
  downstream adaptation via self-evolution: A case study on superglue,''
  \emph{arXiv preprint}, 2022.

\bibitem{yenicelik-etal-2020-bert}
D.~Yenicelik \emph{et~al.}, ``How does {BERT} capture semantics? a closer look
  at polysemous words,'' in \emph{BlackboxNLP Workshop}, 2020.

\bibitem{wu2020slotrefine}
D.~Wu, L.~Ding, F.~Lu, and J.~Xie, ``Slotrefine: A fast non-autoregressive
  model for joint intent detection and slot filling,'' in \emph{EMNLP}, 2020.

\bibitem{wu2021bridging}
D.~Wu \emph{et~al.}, ``Bridging the gap between clean data training and
  real-world inference for spoken language understanding,'' \emph{arXiv
  preprint}, 2021.

\bibitem{cao2021towards}
Y.~Cao, L.~Ding, Z.~Tian, and M.~Fang, ``Towards efficiently diversifying
  dialogue generation via embedding augmentation,'' in \emph{ICASSP}, 2021.

\bibitem{zan2022bridging}
C.~Zan, L.~Ding, L.~Shen, Y.~Cao, W.~Liu, and D.~Tao, ``Bridging cross-lingual
  gaps during leveraging the multilingual sequence-to-sequence pretraining for
  text generation,'' \emph{arXiv preprint}, 2022.

\bibitem{zhang2022bliss}
Z.~Zhang, L.~Ding \emph{et~al.}, ``Bliss: Robust sequence-to-sequence learning
  via self-supervised input representation,'' \emph{arXiv preprint}, 2022.

\bibitem{zhong2022improving}
Q.~Zhong \emph{et~al.}, ``Improving sharpness-aware minimization with fisher
  mask for better generalization on language models,'' in \emph{Findings of
  EMNLP}, 2022.

\bibitem{zhong2022e2s2}
Q.~Zhong, L.~Ding \emph{et~al.}, ``E2s2: Encoding-enhanced sequence-to-sequence
  pretraining for language understanding and generation,'' \emph{arXiv
  preprint}, 2022.

\bibitem{rao2022does}
J.~Rao, F.~Wang, L.~Ding, S.~Qi, Y.~Zhan, W.~Liu, and D.~Tao, ``Where does the
  performance improvement come from? -a reproducibility concern about
  image-text retrieval,'' in \emph{SIGIR}, 2022.

\bibitem{rao2023dynamic}
J.~Rao, L.~Ding, S.~Qi, M.~Fang, Y.~Liu, L.~Shen, and D.~Tao, ``Dynamic
  contrastive distillation for image-text retrieval,'' \emph{TMM}, 2023.

\bibitem{zhong2023knowledge}
Q.~Zhong, L.~Ding, J.~Liu, B.~Du, H.~Jin, and D.~Tao, ``Knowledge graph
  augmented network towards multiview representation learning for aspect-based
  sentiment analysis,'' \emph{TKDE}, 2023.

\bibitem{vegamt}
C.~Zan, K.~Peng \emph{et~al.}, ``Vega-mt: The jd explore academy translation
  system for wmt22,'' \emph{arXiv preprint}, 2022.

\bibitem{zan2022complementarity}
C.~Zan, L.~Ding \emph{et~al.}, ``On the complementarity between pre-training
  and random-initialization for resource-rich machine translation,'' in
  \emph{COLING}, 2022.

\bibitem{2020Fine}
J.~Dodge, G.~Ilharco, R.~Schwartz, A.~Farhadi, H.~Hajishirzi, and N.~Smith,
  ``Fine-tuning pretrained language models: Weight initializations, data
  orders, and early stopping,'' \emph{arXiv preprint}, 2020.

\bibitem{liu2021pre}
P.~Liu, W.~Yuan, J.~Fu, Z.~Jiang, H.~Hayashi, and G.~Neubig, ``Pre-train,
  prompt, and predict: A systematic survey of prompting methods in natural
  language processing,'' \emph{arXiv preprint}, 2021.

\bibitem{zhong2022panda}
Q.~Zhong, L.~Ding \emph{et~al.}, ``Panda: Prompt transfer meets knowledge
  distillation for efficient model adaptation,'' \emph{arXiv preprint}, 2022.

\bibitem{schick2020exploiting}
T.~Schick and H.~Sch{\"u}tze, ``Exploiting cloze questions for few shot text
  classification and natural language inference,'' \emph{arXiv preprint}, 2020.

\bibitem{reynolds2021prompt}
L.~Reynolds and K.~McDonell, ``Prompt programming for large language models:
  Beyond the few-shot paradigm,'' in \emph{Extended Abstracts of the 2021 CHI
  Conference on Human Factors in Computing Systems}, 2021.

\bibitem{levy2017zero}
O.~Levy, M.~Seo, E.~Choi, and L.~Zettlemoyer, ``Zero-shot relation extraction
  via reading comprehension,'' \emph{arXiv preprint}, 2017.

\bibitem{obamuyide-vlachos-2018-zero}
A.~Obamuyide and A.~Vlachos, ``Zero-shot relation classification as textual
  entailment,'' in \emph{FEVER}, 2018.

\end{thebibliography}

\end{document}